\documentclass[10pt,twocolumn]{article}

\usepackage[T1]{fontenc}
\usepackage[utf8]{inputenc}
\usepackage{lmodern}
\usepackage[numbers]{natbib}
\usepackage{pgfplots}
\pgfplotsset{compat=1.18}
\usepackage[
  top=1in,
  bottom=1in,
  left=0.7in,
  right=0.7in
]{geometry}

\usepackage{hyperref}
\usepackage{abstract}
\usepackage{microtype}
\usepackage{graphicx}

\hypersetup{
  colorlinks=true,
  linkcolor=blue,
  urlcolor=blue
}

\begin{document}

\twocolumn[{%
\begin{@twocolumnfalse}

\begin{center}
{\LARGE\bfseries CIFQA: A Deterministic Tool-Grounded Multi-Agent LLM Framework for Financial Query Answering}\\[10pt]

{\large Kunjesh Parekh$^{1*}$ \quad Dr.\ Anil Kumar Tiwari$^{1}$ \quad Dr.\ Divya Saxena$^{1}$}\\[4pt]

{\normalsize $^{1}$School of Artificial Intelligence and Data Science,
Indian Institute of Technology Jodhpur, Rajasthan 342030, India}\\[2pt]

{\small $^{*}$Corresponding author: \href{mailto:P23ai0003@iitj.ac.in}{P23ai0003@iitj.ac.in} \quad
Co-author: \href{mailto:akt@iitj.ac.in}{akt@iitj.ac.in} \quad
Co-author: \href{mailto:divyasaxena@iitj.ac.in}{divyasaxena@iitj.ac.in}}\\[8pt]
\end{center}

\begin{abstract}
Calculation-intensive financial question answering requires not only language understanding, but also exact executable reasoning over structured rates, temporal conditions, numerical formulas, and rule-based constraints. Large Language Models (LLMs), despite strong performance on natural language tasks, often fail on such queries because probabilistic text generation does not consistently execute deterministic multi-step computation. As a result, LLMs may produce arithmetic hallucinations: numerically incorrect but plausible financial answers, even when relevant formulas and contextual information are available. We introduce Calculation-Intensive Financial Query Answering, CIFQA, a deterministic tool-grounded multi-agent LLM framework for financial query answering. CIFQA separates linguistic interpretation from numerical execution by assigning LLM agents to query understanding, routing, parameter extraction, computation planning, and response formulation, while deterministic Python-based tools perform financial operations such as rate lookup, calendar-aware tenure computation, compounding, payout handling, and rule application. In this way, CIFQA treats financial QA as an executable reasoning task rather than a purely text-generation problem. We instantiate CIFQA for fixed deposit query answering and evaluate it on a curated benchmark of fixed deposit queries. CIFQA achieves 95.54\% accuracy on calculation-intensive queries and 90.87\% overall accuracy, substantially outperforming direct LLM baselines even when 
provided with complete formulas, rate cards, and benchmark 
instructions — confirming the limitation is architectural 
rather than informational. Category-wise analysis shows strong performance on quarterly payout, monthly payout, cumulative FD, TDS-related, and edge-case queries. Ablation results further demonstrate that deterministic components, particularly rolling-year adjustment, exact rate lookup, tenure computation, and premature-withdrawal logic, are key contributors to performance. Notably, a 17B open-source backbone operating within CIFQA outperforms substantially larger frontier models including GPT-5.3, Gemini~3, and Claude Sonnet~4.6 evaluated with complete formulas and rate information, demonstrating that architectural design is a more critical determinant of numerical reliability than model scale. While instantiated on fixed deposit queries, CIFQA provides a generalizable and modular design pattern for calculation-intensive financial reasoning, with natural extensions toward multi-instrument decision support, interest rate trend analysis, and cross-asset liquidity optimisation.

\end{abstract}

\vspace{4pt}
\vspace{8pt}

\end{@twocolumnfalse}
}]

\vspace{60pt}

\section{Introduction}

Large Language Models (LLMs) are increasingly being adopted across financial services for customer support, advisory systems, document analysis, and financial question answering \cite{finbert,bloomberggpt,fingpt}. Recent benchmarks such as FinanceBench have further highlighted both the potential and limitations of LLMs for financial-domain question answering \cite{financebench}. Their strong natural language understanding capabilities enable interaction with complex financial information through conversational interfaces. However, financial applications differ fundamentally from conventional text-generation tasks: correctness is often dependent on exact arithmetic execution, calendar-aware reasoning, structured rate retrieval, and strict adherence to financial rules and regulatory conditions. In such settings, even minor numerical deviations can produce financially incorrect outputs, making reliability a critical requirement for real-world deployment \cite{llm_limits_math,cot,pal,gsm_symbolic,numerical_precision_llm,hallucination_survey,factuality_llm}.

\begin{figure*}[!t]
\centering
\includegraphics[width=\textwidth]{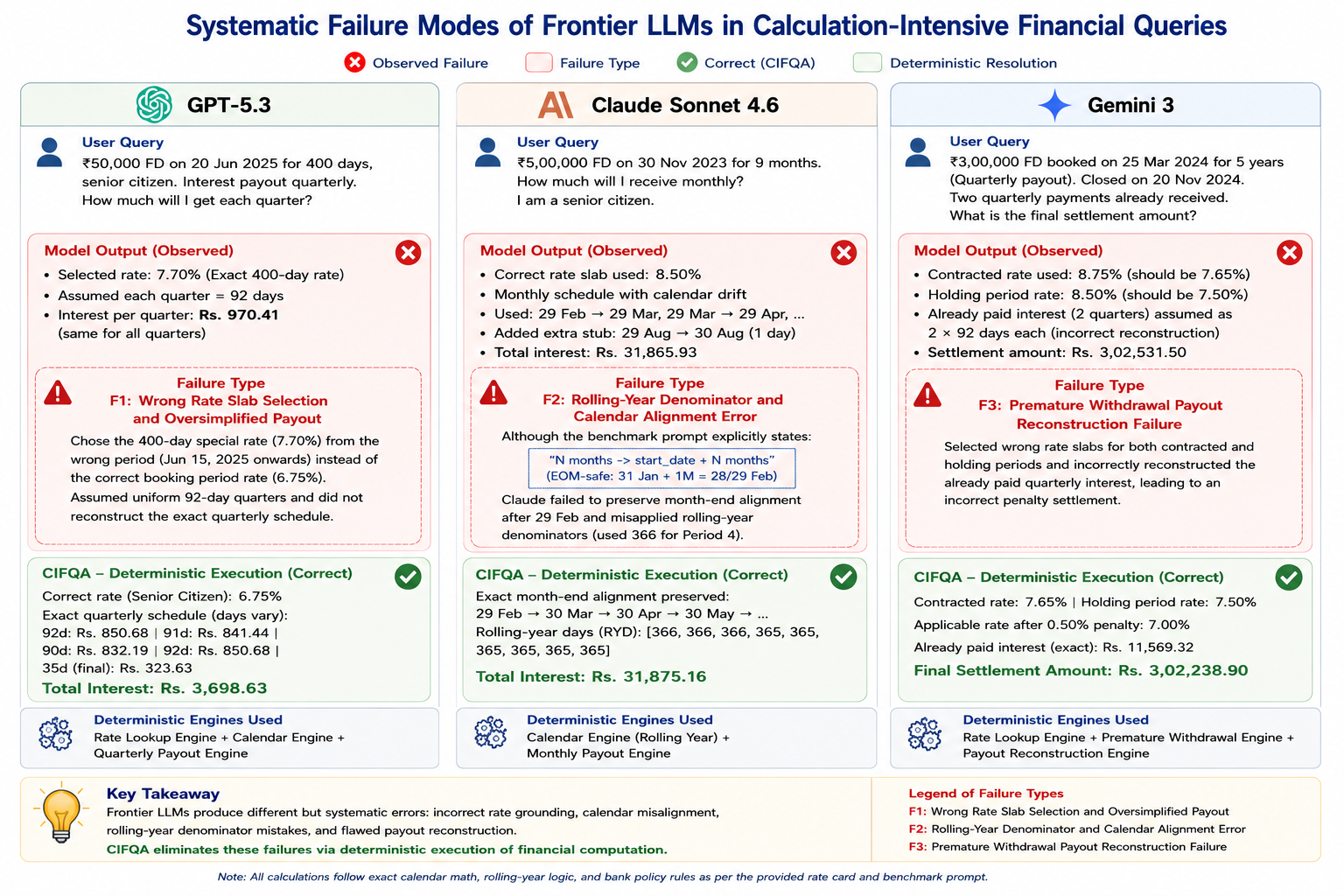}
\caption{Representative execution failures across frontier LLMs on calculation-intensive fixed deposit queries despite complete formulas, rate cards, and benchmark instructions.}
\label{fig:llm_failures}
\end{figure*}

Despite strong language-generation capabilities, frontier LLMs exhibit systematic execution failures on calculation-intensive financial tasks involving exact arithmetic, calendar-aware reasoning, and rule-based financial computation. Figure~\ref{fig:llm_failures} presents representative examples across GPT-5.3, Claude Sonnet 4.6, and Gemini 3 on fixed deposit (FD) queries involving interest-rate selection, rolling-year denominator handling, payout reconstruction, and premature-withdrawal settlement computation.

The observed failures are not isolated numerical mistakes, but systematic execution errors arising from the inability of probabilistic language models to reliably perform deterministic financial computation. Common failure modes include incorrect interest-rate slab selection from structured rate tables, improper handling of rolling-year denominators across leap-year boundaries, payout schedule reconstruction failures, premature-withdrawal settlement inconsistencies, and numerical drift in chained multi-step calculations. Importantly, these failures persist even when models are provided with complete financial formulas, rate cards, and detailed benchmark instructions, indicating that the limitation is architectural rather than informational.

This paper focuses on a class of problems that we define as \textbf{Calculation-Intensive Financial Query Answering (CIFQA)}, where generating a correct response requires exact arithmetic computation over structured financial data under domain-specific financial rules. Representative examples include fixed deposit interest calculation, loan amortization, bond yield estimation, investment return analysis, and tax-related settlement computation \cite{finbert,bloomberggpt,fingpt,financebench}. Unlike conventional financial question answering tasks that primarily involve retrieval, summarization, or textual reasoning, CIFQA requires deterministic execution of computation pipelines involving exact rate grounding, calendar-aware date handling, payout scheduling, compounding, and conditional financial-rule application.

To address this limitation, we propose \textbf{CIFQA}, a deterministic tool-grounded multi-agent LLM framework for executable financial reasoning that builds upon recent advances in tool-augmented reasoning, program-aided language models, and agentic AI systems \cite{pal,toolformer,react,autogen,pot,gorilla,reflexion,llama2,llama3}. CIFQA enforces a strict separation between language understanding and numerical execution: LLM agents are responsible for query interpretation, routing, parameter extraction, computation planning, and response generation, while all arithmetic operations are delegated to deterministic execution engines for rate lookup, calendar-aware computation, payout handling, and financial rule execution. By explicitly preventing LLMs from performing arithmetic operations directly, CIFQA eliminates deterministic computation failures arising from probabilistic token prediction and ensures reproducible financial computation.

We instantiate CIFQA in the domain of fixed deposit (FD) query answering and evaluate it on a curated benchmark of 126 FD queries spanning calculation-intensive reasoning, rate lookup, premature-withdrawal handling, policy interpretation, and edge-case scenarios. CIFQA achieves \textbf{95.54\% accuracy} on calculation-intensive queries and \textbf{90.87\% overall accuracy}, substantially outperforming GPT-5.3, Gemini 3, and Claude Sonnet 4.6. Remarkably, a comparatively smaller 17B open-source backbone operating within CIFQA outperforms substantially larger frontier LLMs evaluated with complete financial formulas, rate cards, and benchmark instructions, demonstrating that deterministic executable computation is more important than model scale for reliable financial reasoning.

The key contributions of this work are as follows:

\begin{itemize}

\item We identify calculation-intensive financial question answering as a distinct class of financial QA problems where correct responses require exact executable reasoning over structured financial data, temporal conditions, numerical formulas, and domain-specific rules. Unlike conventional financial QA tasks that mainly involve retrieval or text generation, these queries require precise operations such as rate selection, calendar-aware computation, compounding, payout handling, and rule application.
\item We propose a deterministic tool-grounded multi-agent LLM framework for financial query answering, CIFQA. The framework separates language-centric tasks from numerical execution: LLM agents perform query interpretation, routing, parameter extraction, computation planning, and response formulation, while deterministic tools execute financial computations and rule-based operations in a verifiable manner.
\item We instantiate the framework for fixed deposit query answering and evaluate it on a curated benchmark of calculation-intensive queries. CIFQA achieves 95.54\% accuracy on calculation-intensive queries and 90.87\% overall accuracy, outperforming direct LLM baselines provided with relevant formulas and rate information. Category-wise and ablation analyses show that deterministic modules such as rolling-year adjustment, exact rate lookup, tenure computation, and premature-withdrawal logic are important for improving calculation correctness and financial reasoning accuracy.




\end{itemize}

The remainder of the paper is organised as follows. Section 2 reviews related work on financial NLP,
retrieval-augmented systems, tool-augmented reasoning,
and multi-agent LLM frameworks. Section~3 formalises the CIFQA problem setting. Section~4 presents the CIFQA framework architecture. Section~5 describes implementation details. Section~6 outlines the experimental setup. Section~7 presents results and analysis. Section~8 discusses generalisation and limitations, and Section~9 concludes the paper.

\section{Related Work}

Research on financial question answering has primarily focused on text understanding, information extraction, sentiment analysis, and financial-domain adaptation of language models \cite{finbert,bloomberggpt,fingpt,financebench,finllm_survey,llm_finance_overview}. Models such as domain-adapted transformers (e.g., FinBERT \cite{finbert}), large-scale financial foundation models such as BloombergGPT \cite{bloomberggpt}, and open financial LLMs such as FinGPT \cite{fingpt} have demonstrated strong performance in financial text understanding, classification, and retrieval tasks. However, these approaches do not address scenarios where responses require precise numerical computation over structured financial data. Financial QA benchmarks such as FinQA \cite{chen2021finqa}, TAT-QA \cite{tatqa}, and ConvFinQA \cite{convfinqa} have demonstrated that numerical reasoning over financial documents remains a significant challenge for existing models, even when domain knowledge is available. 

Retrieval-Augmented Generation (RAG) systems have been widely adopted to improve factual correctness by providing external knowledge to Large Language Models (LLMs) \cite{rag,rag_survey,self_rag}. While effective for knowledge-intensive tasks, RAG frameworks remain insufficient for calculation-intensive settings, as they retrieve information but do not guarantee correct execution of multi-step arithmetic computations.

Recent work has explored chain-of-thought reasoning, tool-augmented inference, program-aided language models, and reasoning-action frameworks, where LLMs generate intermediate reasoning traces, executable programs, or external tool calls to improve numerical reasoning performance \cite{cot,pal,toolformer,react,pot,least_to_most,self_consistency,tot,gorilla,api_bank}. These approaches improve accuracy by augmenting LLMs with external computation capabilities; however, they still rely on the LLM to correctly determine the computation process, generate executable programs, or invoke appropriate tools. Similarly, multi-agent LLM frameworks decompose tasks into specialized roles such as planner, executor, and verifier, improving reasoning through iterative refinement and coordinated agent interactions \cite{autogen,camel,metagpt,multiagent_llm,reflexion,agent_survey,llm_agent_planning_survey}. However, these systems typically retain LLMs within the computation loop, making exact arithmetic correctness dependent on the reliability of intermediate LLM decisions. In contrast, CIFQA removes arithmetic execution entirely from the LLM reasoning loop and delegates all financial computation to deterministic engines, ensuring reproducible and verifiable financial calculations.

Recent research has explored hybrid AI systems that combine language models with deterministic computation, external tools, and executable reasoning modules to improve reliability in structured reasoning tasks \cite{toolformer,pal,react,neurosymbolic,neurosymbolic_review_2025,llm_symbolic_limits}. LLMs have also been shown to exhibit systematic fragility in mathematical reasoning — performance drops significantly when only numerical values in a problem are altered \cite{gsm_symbolic}, and errors accumulate in multi-step arithmetic due to fundamental limitations in numerical precision \cite{numerical_precision_llm,llm_math_errors,dziri_compositionality}. While these approaches improve reasoning accuracy, they generally rely on LLMs to generate executable programs, select tools, or orchestrate computation steps. In contrast, CIFQA enforces a stricter separation between language understanding and numerical execution by removing arithmetic computation entirely from the LLM reasoning loop and delegating all financial calculations to deterministic engines. To the best of our knowledge, deterministic tool-grounded architectures have not been systematically evaluated for calculation-intensive financial query answering tasks involving calendar-aware computation, structured rate lookup, payout scheduling, and financial-rule execution.

In contrast to existing approaches, \textbf{CIFQA} enforces a strict separation between language processing and numerical computation. Instead of relying on LLM-generated reasoning or code, all arithmetic operations are executed through deterministic computation engines. This design eliminates arithmetic hallucinations at their source and ensures exact correctness in calculation-intensive financial queries.

\section{Problem Definition: CIFQA}

We define \textbf{Calculation-Intensive Financial Query Answering (CIFQA)} as a class of question answering tasks in which generating a correct response requires exact numerical computation over structured financial data, along with rule-based reasoning under domain-specific constraints.

Formally, a CIFQA query $q$ consists of a natural language input describing a financial task, which must be mapped to a structured representation containing parameters such as principal amount, interest rate, tenure, compounding frequency, and applicable rules (e.g., taxation thresholds or payout conditions). The goal is to compute an exact numerical output $y$ such that:
\[
y = f(q, R, \mathcal{C}),
\]
where $R$ represents structured financial data (e.g., rate tables), and $\mathcal{C}$ denotes domain-specific computational rules, including calendar-aware calculations, compounding formulas, and regulatory constraints.

Unlike conventional question answering tasks, CIFQA exhibits the following defining characteristics:

\begin{itemize}
    \item \textbf{Exactness requirement:} Even minor numerical deviations render outputs incorrect in real-world financial applications.
    \item \textbf{Multi-step computation:} Queries require chained arithmetic operations, often involving compounding and time-based calculations.
    \item \textbf{Structured data dependency:} Accurate responses depend on correct retrieval of values from rate tables and policy documents.
    \item \textbf{Rule-based logic:} Conditional rules such as tax deductions, penalty clauses, or payout frequencies must be applied correctly.
    \item \textbf{Temporal sensitivity:} Computations often depend on calendar-specific factors such as leap years, day-count conventions, and rolling periods.
\end{itemize}

To evaluate CIFQA systems, we construct a curated evaluation set of 126 fixed deposit (FD) queries designed by domain experts. The evaluation set spans three primary categories: (i) calculation-intensive queries requiring precise interest computation, (ii) interest-rate lookup queries involving retrieval of applicable FD rates under varying customer and tenure conditions, and (iii) policy-related queries involving interpretation of financial rules, taxation conditions, and payout policies. In addition, multiple edge-case scenarios are included within these categories to evaluate robustness under unusual or boundary conditions.

For calculation-intensive queries, ground truth answers are computed manually and validated using spreadsheet-based implementations to ensure exact numerical correctness. Model outputs are evaluated using strict tolerance-based evaluation criteria, requiring both numerical accuracy and correct formatting, reflecting the precision requirements of real-world financial applications.

CIFQA differs fundamentally from existing financial QA approaches \cite{finbert,bloomberggpt,fingpt,financebench} in that it emphasizes deterministic computation rather than textual understanding. While existing financial LLMs and benchmarks primarily focus on financial language understanding, information retrieval, sentiment analysis, or financial-domain knowledge, CIFQA targets exact executable financial reasoning where numerical correctness is the primary evaluation criterion. This distinction necessitates architectures that integrate deterministic computation with LLM-based language understanding, motivating the design of the CIFQA framework.

\section{CIFQA Framework}


The CIFQA framework is designed to eliminate arithmetic hallucinations by enforcing a strict separation between language understanding and numerical computation. The design is inspired by recent advances in tool-augmented reasoning and agentic AI systems \cite{toolformer,pal,react,autogen,camel,metagpt}, but differs in that arithmetic execution is entirely delegated to deterministic computation engines rather than being orchestrated through LLM-generated reasoning steps. Instead of relying on Large Language Models (LLMs) to perform arithmetic reasoning, CIFQA decomposes the problem into specialized components, where LLM-based agents handle interpretation and planning, and deterministic computation engines execute all numerical operations.

\subsection{Framework Overview}


Building upon the failure modes identified in Figure~\ref{fig:llm_failures}, CIFQA addresses the limitations of direct LLM-based financial query answering through a strict separation between language understanding and deterministic computation. Instead of relying on LLMs to perform arithmetic reasoning internally, CIFQA delegates all numerical operations to deterministic execution engines while restricting LLM agents to query interpretation, planning, and response generation.

\begin{figure*}[!t]
\centering
\includegraphics[width=\textwidth]{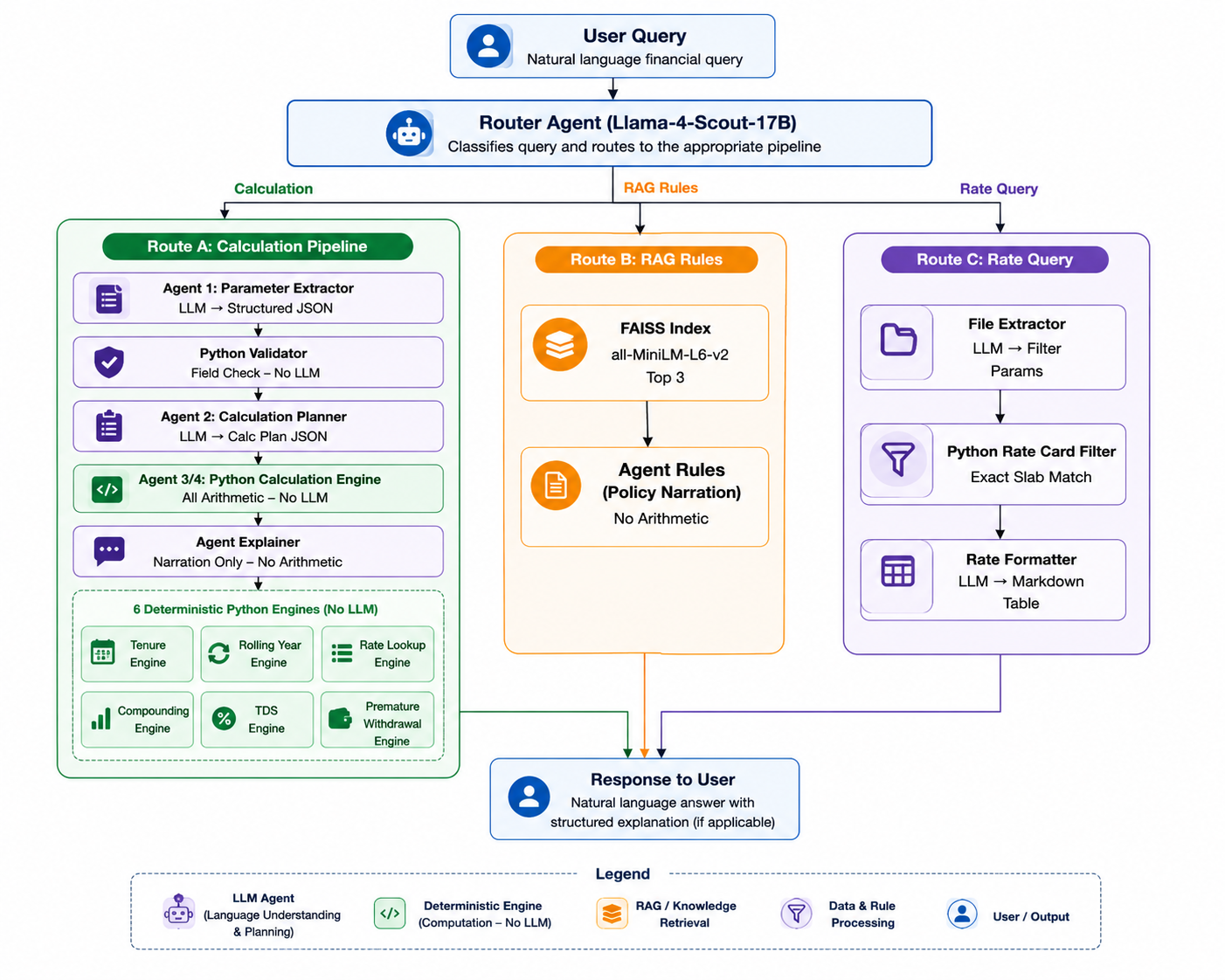}
\caption{CIFQA framework architecture showing separation between LLM-based agents and deterministic computation engines.}
\label{fig:cifqa_architecture}
\end{figure*}

Given a natural language financial query, CIFQA processes the input through a multi-agent LLM pipeline consisting of five stages: (i) routing, (ii) parameter extraction, (iii) planning, (iv) deterministic execution, and (v) response generation. Figure~\ref{fig:cifqa_architecture} illustrates the overall architecture.

At a high level, the framework operates as follows: the input query is first classified to determine whether it requires computation, policy reasoning, or a combination of both. Relevant parameters are then extracted and structured, followed by the generation of a computation plan. All numerical operations are subsequently executed by deterministic computation engines, and the final answer is synthesized into a natural language response.

\subsection{Multi-Agent LLM Design}

CIFQA employs a modular multi-agent LLM architecture, where each agent is responsible for a specific stage of the pipeline. The router, extractor, planner, and response generator collectively form a coordinated multi-agent LLM system, while deterministic computation engines execute all numerical operations outside the LLM reasoning loop.

\paragraph{Router Agent}
The router agent classifies incoming queries into categories such as calculation-intensive, policy-based, or hybrid queries. This step determines whether deterministic computation is required and selects the appropriate execution path.

\paragraph{Extractor Agent}
The extractor agent converts the natural language query into a structured representation by identifying key parameters such as principal amount, tenure, interest rate, payout frequency, and applicable conditions. This step is critical for ensuring that downstream computation receives accurate inputs.

\paragraph{Planner Agent}
The planner agent generates a structured computation plan based on the extracted parameters and domain rules. For calculation-intensive queries, this includes selecting the appropriate formula, determining compounding intervals, handling partial periods, and incorporating calendar-aware adjustments.

\paragraph{Executor (Deterministic Computation Engines)}
All numerical computation is performed by deterministic Python-based computation engines. These include components for rate lookup, interest calculation, calendar-aware day-count computation, and rule-based logic such as tax deduction or payout handling. By delegating all arithmetic operations to these engines, CIFQA ensures exact numerical correctness.

\paragraph{Response Generator}
The response generator converts the computed outputs into a coherent natural language response. This includes formatting numerical values, incorporating explanatory steps if required, and ensuring clarity for end users.

\subsection{Separation of Language Understanding and Computation}

A key design principle of CIFQA is the strict separation between language understanding and deterministic computation, building upon prior work in tool-grounded and program-aided reasoning while enforcing complete removal of arithmetic execution from the LLM loop \cite{toolformer,pal,react}. LLMs are never used to perform arithmetic operations; instead, they are limited to language understanding, planning, and response synthesis. All computations are executed by deterministic engines, ensuring reproducibility and eliminating numerical drift.

This separation addresses a fundamental limitation of LLMs: while they are effective at interpreting language, they are inherently unreliable for exact numerical computation due to their probabilistic nature. By removing arithmetic reasoning from the LLM pipeline, CIFQA eliminates arithmetic hallucinations at their source.

\subsection{Execution Flow}

\begin{figure*}[!t]
\centering
\includegraphics[width=\textwidth]{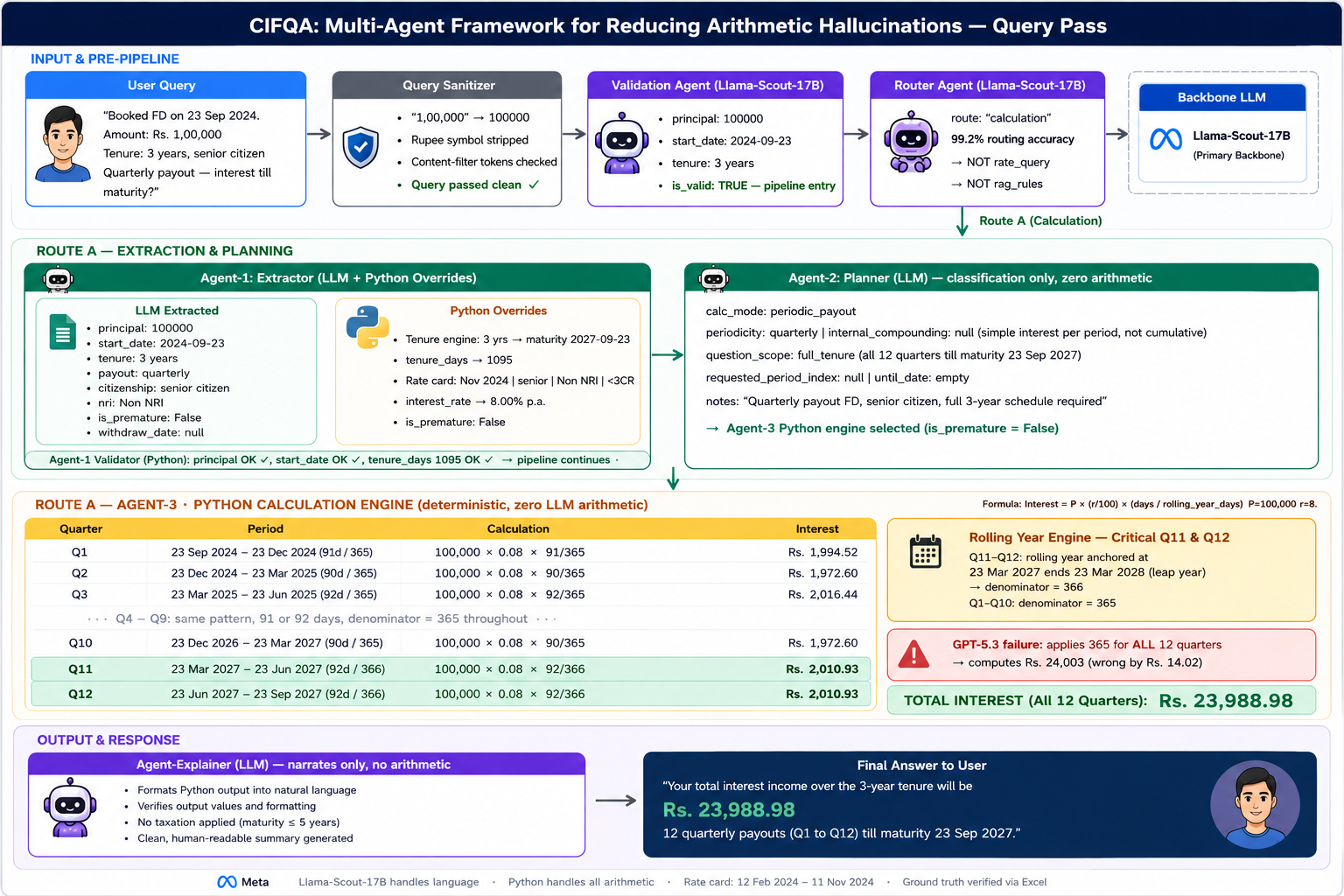}
\caption{Example end-to-end execution flow of CIFQA for a calculation-intensive fixed deposit query. The figure illustrates query sanitization, validation, routing, parameter extraction, planning, deterministic computation, and final response generation. All arithmetic operations are executed through deterministic computation engines, while LLM agents are restricted to language understanding, planning, and response synthesis.}
\label{fig:query_flow}
\end{figure*}

The end-to-end execution flow of CIFQA is illustrated in Figure~\ref{fig:query_flow} and can be summarized as follows:

\begin{enumerate}
    \item A user submits a natural language financial query.
    \item The router agent determines the query type and execution path.
    \item The extractor agent converts the query into structured parameters.
    \item The planner agent generates a computation plan.
    \item Deterministic computation engines execute all numerical operations.
    \item The response generator produces the final answer.
\end{enumerate}

This pipeline ensures that all stages of computation are transparent, modular, and verifiable, making CIFQA suitable for deployment in high-stakes financial applications where correctness is critical.

\section{Implementation}

The CIFQA framework is implemented as a hybrid multi-agent LLM system combining Large Language Model (LLM) inference with deterministic Python-based computation engines. The implementation is designed to ensure reproducibility, modularity, and exact numerical correctness.

\subsection{LLM Backend}

All language understanding and planning tasks are performed using an LLM accessed via API-based inference. The model is responsible for query routing, parameter extraction, computation planning, and response generation. Importantly, the LLM is not used for any arithmetic computation, ensuring that all numerical operations remain deterministic.

\subsection{Deterministic Computation Engines}

Numerical computation is handled by a set of modular Python-based engines, each responsible for a specific aspect of financial calculation. These include:

\begin{itemize}
    \item \textbf{Rate Lookup Engine:} Retrieves applicable interest rates from structured rate tables based on tenure and customer category.
    \item \textbf{Interest Computation Engine:} Performs exact interest calculations using appropriate compounding formulas and payout schedules.
    \item \textbf{Calendar Engine:} Handles date-related computations, including leap years, day-count conventions, and rolling periods.
    \item \textbf{Rule Engine:} Applies conditional financial rules such as tax deduction thresholds and payout-specific adjustments.
\end{itemize}

All engines are implemented using deterministic logic to ensure consistency and reproducibility across executions.

\subsection{Knowledge Sources}

The framework utilizes structured financial data sources, including rate tables and policy rules, which are incorporated into the system as deterministic lookup modules. This eliminates ambiguity in parameter selection and ensures alignment with real-world financial specifications.

\subsection{System Integration}

The CIFQA pipeline is orchestrated through a sequential execution flow, where outputs from each agent are passed as structured inputs to subsequent components. The system is implemented in Python, with clear separation between LLM-driven components and deterministic computation modules.

This modular design allows individual components to be independently updated or extended, enabling adaptation to other financial domains without modifying the overall architecture.

\section{Experimental Setup}

This section describes the evaluation protocol used to assess the performance of CIFQA on calculation-intensive financial queries.

\subsection{Evaluation Dataset}

We evaluate CIFQA on a curated set of 126 fixed deposit (FD) queries designed by domain experts. The evaluation set covers three primary categories:

\begin{itemize}
    \item \textbf{Calculation-intensive queries:} Require precise interest computation involving compounding, partial periods, and calendar-aware calculations.
    
    \item \textbf{Interest-rate lookup queries:} Require retrieval of applicable FD interest rates under varying tenure, customer category, deposit amount, and booking-date conditions.
    
    \item \textbf{Policy-related queries:} Involve interpretation of financial rules such as taxation thresholds, payout conditions, premature withdrawal policies, and TDS-related scenarios.
\end{itemize}

In addition, multiple edge-case scenarios are included across these categories to evaluate robustness under unusual or boundary conditions, including irregular tenures, large principal amounts, ambiguous phrasing, and calendar-sensitive calculations.

For the subset of 101 calculation-intensive queries, ground truth answers are computed manually and verified using spreadsheet-based implementations to ensure exact numerical correctness.

\subsection{Evaluation Metrics}

We adopt strict evaluation criteria reflecting the requirements of real-world financial applications. A response is considered correct only if:

\begin{itemize}
    \item The final numerical value matches the ground-truth value within an absolute tolerance of \(\pm 1\) INR.
    
    \item Intermediate computations (where applicable) follow correct financial logic.
    
    \item The output formatting is consistent with expected financial representations.
\end{itemize}

Any response producing a numerical deviation beyond the accepted tolerance threshold is classified as an \textbf{arithmetic hallucination}. This includes errors arising from incorrect arithmetic execution, improper day-count handling, incorrect rate application, compounding mistakes, or failure to apply domain-specific financial rules correctly.

This strict evaluation protocol ensures that numerically meaningful deviations are treated as errors, distinguishing CIFQA from conventional QA benchmarks that allow approximate semantic correctness.

\subsection{Baseline Models}

We compare CIFQA against several state-of-the-art frontier Large Language Models (LLMs), including:

\begin{itemize}
    \item GPT-5.3 (OpenAI)
    \item Gemini 3 (Google)
    \item Claude Sonnet 4.6 (Anthropic)
\end{itemize}

This evaluation setting is intentionally designed to provide frontier LLMs with complete financial knowledge and computational instructions, allowing the study to isolate failures arising from numerical execution rather than information retrieval or missing domain knowledge.

To ensure a fair comparison against production-grade systems, all frontier LLM baselines were evaluated through their native conversational interfaces rather than raw API inference. This setup allows the models to utilize their full deployed capabilities, including tool access, enhanced reasoning orchestration, and interface-level optimizations available to end users. Preliminary experiments using direct API-only inference resulted in substantially lower and less stable numerical accuracy across models. Therefore, reported baseline results reflect the strongest practically deployable performance of each system.

\subsection{Evaluation Protocol}

Each query is independently evaluated by generating responses from baseline models and CIFQA. Model outputs are manually compared against ground truth answers to assess correctness. For calculation-intensive queries, evaluation focuses on exact numerical accuracy, while policy and edge-case queries are evaluated based on logical correctness and adherence to domain rules.

This evaluation protocol ensures a fair and rigorous comparison between CIFQA and baseline LLMs, particularly in scenarios requiring precise numerical reasoning.

\section{Results and Analysis}

This section evaluates CIFQA with a primary focus on its ability to eliminate arithmetic hallucinations in calculation-intensive financial queries. The evaluation explicitly distinguishes between direct LLM inference and structured multi-agent LLM execution, highlighting the role of architectural design in achieving numerical correctness.

\subsection{Evaluation Setup and Comparison Protocol}

Baseline models (GPT-5.3, Gemini 3, and Claude Sonnet) are evaluated using a single-prompt setup, where each query is provided along with relevant contextual information such as formulas and rate tables. The models are required to directly generate the final answer without intermediate structured reasoning or external computation.

In contrast, CIFQA follows a multi-agent LLM execution paradigm, where the underlying LLM operates within a structured pipeline consisting of routing, parameter extraction, planning, and deterministic execution. The CIFQA framework is instantiated using Llama-Scout-17B as the primary backbone model.

To isolate the effect of architectural design from model scale, the same CIFQA pipeline is evaluated with alternative backbone models, including Llama-70B and Llama-8B, while keeping all other components unchanged. This enables a controlled comparison between model capacity and system design.

The CIFQA system is tuned using a small set of 21 development queries to refine prompts and ensure stable agent behavior. All reported results are evaluated on the remaining queries, ensuring that performance reflects generalization rather than memorization.

\subsection{Calculation-Intensive Performance}

Table~\ref{tab:calc_results} presents performance on 101 calculation-intensive queries.

\begin{table}[h]
\centering
\caption{Performance on calculation-intensive queries}
\label{tab:calc_results}
\begin{tabular}{l c}
\hline
\textbf{Model} & \textbf{Accuracy (\%)} \\
\hline
GPT-5.3 & 45.05 \\
Gemini 3 & 70.30 \\
Claude Sonnet 4.6 & 83.66 \\
CIFQA (Llama-70B backbone) & 89.60 \\
CIFQA (Llama-8B backbone) & 68.81 \\
\textbf{CIFQA (Llama-Scout-17B)} & \textbf{95.54} \\
\hline
\end{tabular}
\end{table}

\begin{figure}[h]
\centering
\begin{tikzpicture}
\begin{axis}[
    ybar,
    ymin=0,
    ymax=105,
    enlarge x limits=0.12,
    ylabel={Accuracy (\%)},
    symbolic x coords={GPT-5.3, Gemini 3, Claude Sonnet 4.6, Llama-8B, Llama-70B, CIFQA},
    xtick=data,
    xticklabel style={rotate=40, anchor=east,font=\small},
    nodes near coords,
    nodes near coords style={font=\small},
    bar width=8pt,
    width=0.95\columnwidth,
]
\addplot coordinates {
    (GPT-5.3,45.05)
    (Gemini 3,70.30)
    (Claude Sonnet 4.6,83.66)
    (Llama-8B,68.81)
    (Llama-70B,89.60)
    (CIFQA,95.54)
};
\end{axis}
\end{tikzpicture}
\caption{Accuracy comparison on calculation-intensive queries}
\label{fig:calc_comparison}
\end{figure}
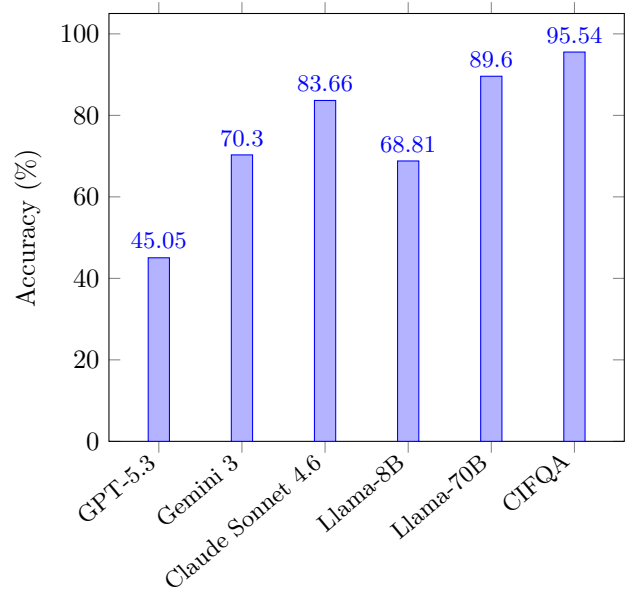

CIFQA achieves \textbf{95.54\% accuracy}, significantly outperforming both proprietary frontier models and larger open-source baselines. This improvement is achieved despite baseline models operating in a direct single-prompt setting, while CIFQA leverages a structured multi-agent LLM execution framework.

The results demonstrate that increasing model scale alone is insufficient to resolve arithmetic unreliability in financial reasoning. While larger models such as Llama-70B improve performance compared to smaller models, they remain inferior to CIFQA. This confirms that the primary limitation of LLMs in financial reasoning lies in unreliable arithmetic execution rather than insufficient model capacity.

\subsection{Arithmetic Hallucination Rate}

To directly evaluate numerical reliability, we additionally report the \textbf{Arithmetic Hallucination Rate (AHR)}, defined as the percentage of calculation-intensive queries producing numerically incorrect outputs beyond the accepted evaluation tolerance.

Formally,

\[
\mathrm{AHR}(\%) = 100 - \mathrm{Accuracy}(\%)
\]

where accuracy is measured under the strict financial correctness criteria defined in Section~6.

Table~\ref{tab:ahr_results} presents the corresponding arithmetic hallucination rates for all evaluated models.

\begin{table}[h]
\centering
\caption{Arithmetic Hallucination Rate (AHR) on calculation-intensive queries}
\label{tab:ahr_results}
\begin{tabular}{l c}
\hline
\textbf{Model} & \textbf{AHR (\%)} \\
\hline
GPT-5.3 & 54.95 \\
Gemini 3 & 29.70 \\
Claude Sonnet 4.6 & 16.34 \\
CIFQA (Llama-70B backbone) & 10.40 \\
CIFQA (Llama-8B backbone) & 31.19 \\
\textbf{CIFQA (Llama-Scout-17B)} & \textbf{4.46} \\
\hline
\end{tabular}
\end{table}

The results show that CIFQA substantially reduces arithmetic hallucinations compared to both proprietary and open-source LLM baselines. The low hallucination rate achieved by CIFQA demonstrates that deterministic computation is highly effective for calculation-intensive financial reasoning tasks.

\subsection{Category-wise Analysis}

\begin{table*}[!t]
\centering
\caption{Category-wise accuracy comparison across CIFQA and frontier LLMs}
\label{tab:category_results}

\resizebox{\textwidth}{!}{
\begin{tabular}{lcccc}
\hline
\textbf{Query Category} &
\textbf{CIFQA} &
\textbf{Claude Sonnet 4.6} &
\textbf{Gemini 3} &
\textbf{GPT-5.3} \\
\hline

\multicolumn{5}{c}{\textbf{Calculation-Intensive Categories}} \\
\hline

Quarterly Payout & 100.00\% & 92.00\% & 80.00\% & 44.00\% \\
TDS / Mismatch & 100.00\% & 100.00\% & 100.00\% & 25.00\% \\
Cumulative FD & 100.00\% & 71.43\% & 71.43\% & 57.14\% \\
Monthly Payout & 100.00\% & 81.82\% & 68.18\% & 59.09\% \\
Premature Withdrawal & 92.86\% & 78.57\% & 35.71\% & 25.00\% \\
Half-Yearly Payout & 100.00\% & 71.43\% & 100.00\% & 71.43\% \\
Rate Analytics Query & 58.33\% & 75.00\% & 83.33\% & 66.67\% \\
Edge Cases & 100.00\% & 75.00\% & 68.75\% & 50.00\% \\
Robustness / Invalid & 87.50\% & 100.00\% & 37.50\% & 18.75\% \\

\hline
\multicolumn{5}{c}{\textbf{Retrieval / Policy-Oriented Category}} \\
\hline

RAG Rules & 72.00\% & 100.00\% & 92.00\% & 88.00\% \\

\hline
Overall Accuracy & 90.87\% & 86.90\% & 74.60\% & 53.57\% \\
Calculation-Intensive Queries & \textbf{95.54\%} & 83.66\% & 70.30\% & 45.05\% \\
\hline
\end{tabular}
}
\end{table*}

Table~\ref{tab:category_results} presents a category-wise comparison between CIFQA and frontier LLMs. CIFQA achieves near-perfect performance across most calculation-intensive categories, including quarterly payout, monthly payout, cumulative FD, TDS-related queries, edge cases, and half-yearly payout scenarios. The largest performance gains relative to frontier LLMs are observed in premature-withdrawal and payout-based calculations, which require precise calendar-aware execution, exact rate grounding, and rule-based financial computation.

The category-wise results further demonstrate that deterministic execution is particularly beneficial for financial tasks involving multi-step arithmetic reasoning. Across the calculation-intensive categories, CIFQA consistently outperforms all frontier LLM baselines, achieving perfect accuracy in several categories where even the strongest baseline models exhibit substantial performance degradation. These findings indicate that the dominant source of error in financial reasoning systems is not language understanding, but unreliable numerical execution and rule application.

The results further reveal that not all categories within the benchmark rely equally on deterministic arithmetic execution. While CIFQA substantially outperforms all baseline models across traditional calculation-intensive categories, its performance is comparatively lower on the \emph{Rate Analytics Query} and \emph{RAG Rules} categories. Importantly, this does not reflect a limitation in rate retrieval, as CIFQA performs exact rate lookup deterministically. Rather, Rate Analytics queries require higher-level reasoning over structured rate tables to identify patterns, relationships, and trends across customer segments, deposit amounts, and tenure ranges. Similarly, RAG Rules queries depend more heavily on policy interpretation and retrieval than numerical execution. This behavior is consistent with the design objective of CIFQA, which prioritizes reliable financial computation through deterministic execution while providing limited optimization for analytical reasoning and retrieval-intensive policy interpretation.

\subsection{Overall Performance}

For completeness, CIFQA achieves \textbf{90.87\% overall accuracy}. The gap between calculation accuracy and overall accuracy is primarily driven by policy-heavy queries, which are outside the primary optimization scope of CIFQA.

\subsection{Error Analysis}

Baseline LLMs exhibit consistent failure modes in calculation-intensive queries, including incorrect compounding, misinterpretation of tenure, and numerical drift in multi-step reasoning.

We observe three dominant arithmetic failure categories across frontier LLMs:

\begin{itemize}
    \item \textbf{F1: Structured parameter misselection} — incorrect retrieval or application of interest-rate slabs, payout conditions, or tenure mappings.
    
    \item \textbf{F2: Calendar-aware execution errors} — incorrect handling of rolling-year computations, leap years, quarterly boundaries, and date-sensitive interest schedules.
    
    \item \textbf{F3: Numerical precision drift} — accumulation of rounding inconsistencies and arithmetic deviations during multi-step financial calculations.
\end{itemize}

Figure~\ref{fig:llm_failures} presents representative examples of these failure categories across GPT-5.3, Claude Sonnet 4.6, and Gemini 3. The observed errors demonstrate that arithmetic failures are not isolated numerical mistakes but systematic execution failures arising from incorrect parameter grounding, calendar-aware computation errors, and numerical drift during multi-step financial reasoning.

The category-level performance trends shown in Table~\ref{tab:category_results} are consistent with the representative failure cases illustrated in Figure~\ref{fig:llm_failures}. Across frontier LLMs, the dominant sources of error arise from incorrect rate selection, calendar-aware execution failures, payout reconstruction errors, and numerical drift in multi-step financial calculations. CIFQA avoids these failures by delegating all numerical operations to deterministic computation engines while restricting LLM agents to language understanding, planning, and response generation.

CIFQA eliminates these errors by delegating all computation to deterministic computation engines. The remaining errors are primarily associated with:

\begin{itemize}
    \item Ambiguities in policy interpretation for RAG-based queries
    \item Occasional extraction or routing errors
\end{itemize}

\paragraph{Routing Accuracy}
The router component demonstrates near-perfect performance for calculation-intensive queries, with only a single observed misclassification. This indicates that query-type identification is not a major source of error for arithmetic tasks. In contrast, most routing-related errors occur in policy-heavy (RAG-based) queries, where boundaries between retrieval and computation are less clearly defined.

These findings confirm that arithmetic hallucination is the dominant source of error in financial query answering, and that CIFQA effectively addresses this limitation.

\subsection{Ablation Study}
Unlike end-to-end LLM systems, CIFQA enables component-level attribution of numerical reliability through deterministic module isolation.
To isolate the contribution of individual deterministic components, we perform a targeted ablation study by selectively disabling specific computation modules within CIFQA and evaluating their impact on calculation-intensive queries.

\begin{table}[h]
\centering
\caption{Component-level ablation analysis}
\label{tab:ablation}
\resizebox{\columnwidth}{!}{
\begin{tabular}{l c c c}
\hline
\textbf{Ablation Component} & \textbf{Applicable Queries} & \textbf{Correct} & \textbf{Accuracy (\%)} \\
\hline
Tenure Computation Logic & 101 & 85 & 84.65 \\
Rolling-Year Adjustment Logic & 25 & 12 & 48.00 \\
Exact Day-Count ($n$) Computation & 9 & 8 & 88.89 \\
Rate Lookup Module & 101 & 77 & 76.73 \\
Premature Withdrawal Override Logic & 13 & 10 & 76.92 \\
\hline
\end{tabular}
}
\end{table}

Each ablation corresponds to disabling a specific deterministic computation capability within CIFQA while keeping all other components unchanged.

\begin{itemize}
    \item \textbf{Rolling-year adjustment logic} has the largest impact, with accuracy dropping to \textbf{48.00\%}. This highlights the importance of correctly handling multi-period interest calculations across year boundaries, particularly in scenarios involving leap years and partial-year segmentation.

    \item \textbf{Rate lookup module} and \textbf{premature withdrawal override logic} significantly affect performance, reducing accuracy to approximately \textbf{76–77\%}. These components ensure correct parameter grounding and conditional rule application, both of which are essential in real-world financial computations.

    \item \textbf{Tenure computation logic}, which governs segmentation of total duration into compounding intervals, affects all queries and leads to a drop to \textbf{84.65\%}, indicating its role in maintaining structural correctness of calculations.

    \item \textbf{Exact day-count ($n$) computation} has a comparatively localized impact, with accuracy remaining at \textbf{88.89\%}, suggesting that its contribution is limited to specific edge cases rather than the majority of queries.
\end{itemize}

Overall, the ablation study demonstrates that CIFQA's performance is driven by precise deterministic handling of financial computation primitives. In particular, calendar-aware logic and structured parameter grounding emerge as the most critical components, reinforcing the importance of deterministic execution in calculation-intensive financial reasoning.

\section{Discussion}

The results demonstrate that CIFQA effectively eliminates arithmetic hallucinations in calculation-intensive financial queries by enforcing a strict separation between language understanding and numerical computation. This architectural design has broader implications beyond the specific fixed deposit (FD) use case evaluated in this work.

\subsection{Generalization to Other Financial Domains}

\begin{figure}[h]
\centering
\includegraphics[width=\columnwidth]{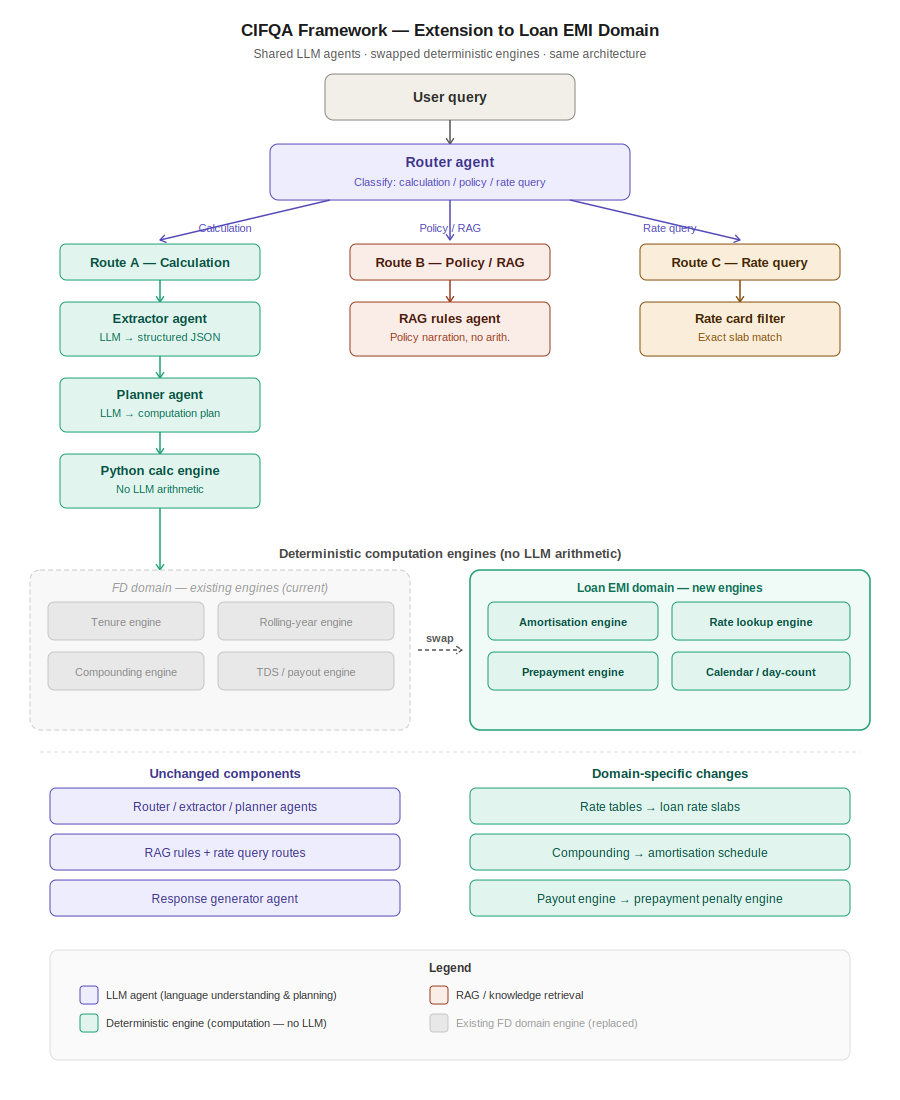}
\caption{CIFQA framework extension to loan EMI domain. Multi-agent LLM components (router, extractor, planner, response generator) remain unchanged across domains. Only the deterministic computation engines are replaced with domain-specific equivalents, confirming the modular and domain-agnostic design of CIFQA.}
\label{fig:emi_extension}
\end{figure}

Although CIFQA is instantiated on FD query answering, the framework is inherently domain-agnostic. The deterministic computation layer can be extended to other financial applications by replacing only domain-specific modules while keeping all multi-agent LLM components unchanged. Figure~\ref{fig:emi_extension} illustrates this concretely for loan EMI calculation: the router, extractor, planner, and response generator agents require no modification, while only four deterministic modules are swapped — the compounding engine is replaced by an amortisation schedule engine, the rolling-year engine by a calendar day-count engine, the payout engine by a prepayment penalty engine, and rate tables are updated to loan rate slabs. This modular boundary is precisely what makes CIFQA domain-agnostic.

Potential applications include:

\begin{itemize}
    \item \textbf{Loan amortization and EMI calculations} — replace compounding and payout engines with amortisation schedule and prepayment penalty engines.
    \item \textbf{Bond pricing and yield computation} — replace rate lookup with yield curve interpolation and day-count convention engines.
    \item \textbf{Portfolio return and investment analysis} — replace interest engines with return computation and rebalancing rule engines.
    \item \textbf{Tax computation and compliance systems} — replace FD rule engine with jurisdiction-specific tax slab and deduction engines.
\end{itemize}

In each case, the core principle remains unchanged: LLMs handle interpretation and planning, while deterministic engines ensure exact numerical correctness.

\subsection{Architecture vs Model Scaling}

A key finding of this work is that increasing model size alone does not resolve arithmetic reasoning errors. Even large-scale models exhibit significant performance gaps in calculation-intensive tasks. In contrast, CIFQA enables smaller models to outperform larger ones by removing the burden of arithmetic reasoning from the LLM.

This suggests that architectural design, rather than model scale, is the primary determinant of performance in domains requiring exact computation. Notably, the 17B CIFQA instantiation outperforms substantially larger proprietary models despite having significantly fewer parameters, indicating that deterministic execution can compensate for limitations in model scale when exact numerical correctness is required. The findings are consistent with prior observations that LLMs often struggle with arithmetic and symbolic reasoning despite scaling improvements \cite{cot,pal,toolformer,gsm_symbolic,llm_symbolic_limits,emergent_abilities,scaling_laws}, and indicate that deterministic execution remains necessary for reliable financial computation.

\subsection{Separation of Concerns in LLM Systems}

CIFQA exemplifies a broader design paradigm for LLM-based systems: separating probabilistic language understanding from deterministic computation through external execution engines \cite{toolformer,pal,react,neurosymbolic,neurosymbolic_review_2025}. This separation improves reliability, interpretability, and reproducibility, particularly in high-stakes domains such as finance.

The results indicate that LLMs are well-suited for tasks involving ambiguity, interpretation, and planning, but should not be relied upon for exact numerical execution. Integrating deterministic computation components provides a principled solution to this limitation.

\subsection{Limitations}

While CIFQA achieves near-perfect performance on calculation-intensive queries, certain limitations remain:

\begin{itemize}
    \item Performance on policy-heavy (RAG-based) queries is comparatively lower, as these tasks involve complex rule interpretation beyond deterministic computation.
    \item The framework relies on accurate parameter extraction and routing, and errors in these stages can propagate to downstream components.
    \item The current evaluation focuses on fixed deposit scenarios, and broader validation across multiple financial domains remains future work.
\end{itemize}

Addressing these limitations will require improved integration of retrieval and reasoning mechanisms alongside deterministic computation.

\subsection{Implications for Financial AI Systems}

The findings of this work suggest that reliable financial AI systems should not rely solely on end-to-end LLM reasoning. Instead, hybrid architectures that combine LLM-based reasoning with deterministic computation offer a more robust and scalable approach.

By ensuring exact numerical correctness, CIFQA provides a pathway for deploying LLM-based systems in real-world financial applications where precision is critical.

\section{Conclusion}

This paper introduced CIFQA, a deterministic tool-grounded 
multi-agent LLM framework for calculation-intensive financial query 
answering. CIFQA addresses a fundamental limitation of contemporary 
LLMs: their inability to reliably execute exact financial 
computations despite strong natural language understanding 
capabilities. By enforcing a strict separation between language 
understanding and numerical execution, CIFQA transforms financial 
query answering from a text-generation problem into an executable 
reasoning problem.

Evaluated on a curated benchmark of fixed deposit queries, CIFQA 
achieves \textbf{95.54\% accuracy on calculation-intensive tasks} 
and \textbf{90.87\% overall accuracy}, significantly outperforming 
both proprietary and open-source LLMs including GPT-5.3, Gemini 3, 
and Claude Sonnet 4.6. Crucially, a 17B open-source backbone 
operating within the CIFQA framework outperforms substantially 
larger frontier models evaluated with complete formulas and rate 
information, demonstrating that architectural design — specifically, 
the strict separation of language understanding from deterministic 
computation — is a more important determinant of numerical 
reliability than model scale.

A detailed ablation study confirms that deterministic components, 
particularly calendar-aware rolling-year computation and exact rate 
lookup, are the primary drivers of this performance advantage. 
These findings highlight that arithmetic hallucinations in financial 
reasoning are an architectural problem, not an information problem, 
and that hybrid systems combining LLM-based interpretation with 
verifiable deterministic execution provide a principled solution.

Beyond the fixed deposit domain, CIFQA establishes a generalizable 
design pattern for calculation-intensive reasoning across financial 
applications. The modular separation between LLM agents and 
deterministic engines means that extending CIFQA to loan 
amortization, bond yield computation, or tax settlement requires 
only replacing domain-specific computation modules, with all 
language understanding components remaining unchanged.

Looking ahead, the most significant opportunity for future work 
lies in extending CIFQA from reactive query answering toward 
proactive financial decision support. Several high-value scenarios 
remain unaddressed by the current system. 
First, \textbf{interest rate trend and pattern analysis}: while 
CIFQA already performs exact rate lookup and identifies the tenure 
bracket yielding maximum interest for a given customer profile, 
a decision-support extension could go further by analysing 
structural patterns across the rate card — for example, 
identifying that senior citizens consistently receive a 0.50 
percentage point premium over regular citizens across all tenure 
slabs, detecting tenure ranges where the rate curve flattens or 
spikes, or flagging booking-period windows where rates have 
historically been most favourable. Such pattern-level reasoning 
requires moving beyond deterministic lookup toward analytical 
inference over structured rate data, representing a natural 
next frontier for the CIFQA framework.
 Second, 
\textbf{multi-FD liquidation planning}: when a customer requires 
urgent liquidity across a portfolio of fixed deposits with 
different tenures, rates, and penalty conditions, the system 
could deterministically compute the net cost of breaking each 
FD — accounting for premature withdrawal penalties, accrued 
interest forfeiture, and residual tenure — and recommend the 
optimal liquidation sequence that minimises financial loss. Third, 
\textbf{cross-asset class optimisation}: in scenarios where a 
customer holds an FD and requires short-term liquidity, the system 
could compare the effective cost of breaking the FD against the 
cost of taking a short-term loan against it. For example, if a 
customer holds an FD earning 7.5\% annually and requires funds for 
30 days, taking a loan at 9\% for one month may be financially 
superior to premature FD closure, since the effective loan cost 
for 30 days is approximately 0.74\% while the penalty and interest 
forfeiture on breaking the FD may exceed 1--2\%. Such 
cross-instrument comparisons require exactly the kind of 
deterministic multi-step computation that CIFQA is architecturally 
designed to support, making this a natural and high-impact 
direction for extension.

These extensions would require augmenting the planner agent with 
multi-objective reasoning capabilities — comparing outcomes across 
instruments, time horizons, and penalty structures — while 
retaining the deterministic execution layer that ensures 
correctness. The CIFQA architecture is well-positioned for this 
evolution: the same separation of concerns that eliminates 
arithmetic hallucinations in single-query answering also provides 
the computational reliability necessary for trustworthy financial 
decision support in multi-instrument, multi-objective scenarios.

\bibliography{references}

@article{finbert,
  title   = {FinBERT: Financial Sentiment Analysis with Pre-trained Language Models},
  author  = {Yang, Yi and others},
  journal = {arXiv preprint arXiv:2006.08097},
  year    = {2020}
}

@article{bloomberggpt,
  title   = {BloombergGPT: A Large Language Model for Finance},
  author  = {Wu, Shijie and Irsoy, Ozan and Lu, Steven and Dabravolski, Vadim and
             Dredze, Mark and Gehrmann, Sebastian and Kambadur, Prabhanjan and
             Rosenberg, David and Mann, Gideon},
  journal = {arXiv preprint arXiv:2303.17564},
  year    = {2023}
}

@article{fingpt,
  title   = {FinGPT: Open-Source Financial Large Language Models},
  author  = {Yang, Hongyang and Liu, Xiao-Yang and Wang, Christina Dan},
  journal = {arXiv preprint arXiv:2306.06031},
  year    = {2023}
}

@article{finllm_survey,
  title   = {A Survey of Large Language Models in Finance (FinLLMs)},
  author  = {Lee, Yupeng and Chen, Yuqi and Huang, Zhijie and Chen, Yifei and
             Liu, Xiaolong},
  journal = {arXiv preprint arXiv:2402.02315},
  year    = {2024}
}

@article{llm_finance_overview,
  title   = {Revolutionizing Finance with LLMs: An Overview of Applications and Insights},
  author  = {Zhao, Huaqin and Liu, Zhengliang and Wu, Zihao and Li, Yiwei and
             Yang, Tianze and Shu, Peng and Xu, Shaochen and Dai, Haixing and
             Zhao, Lin and Mai, Gengchen and Liu, Ninghao and Liu, Tianming},
  journal = {arXiv preprint arXiv:2401.11641},
  year    = {2024}
}

@article{financebench,
  title   = {FinanceBench: A New Benchmark for Financial Question Answering},
  author  = {Islam, Pranab and Kannappan, Anand and Kiela, Douwe and Qian, Rebecca
             and Scherrer, Nino and Vidgen, Bertie},
  journal = {arXiv preprint arXiv:2311.11944},
  year    = {2023}
}

@inproceedings{tatqa,
  title     = {{TAT-QA}: A Question Answering Benchmark on a Hybrid of Tabular and
               Textual Content in Finance},
  author    = {Zhu, Fengbin and Lei, Wenqiang and Huang, Youcheng and Wang, Chao
               and Zhang, Shuo and Lv, Jiancheng and Feng, Fuli and Chua, Tat-Seng},
  booktitle = {Proceedings of the 59th Annual Meeting of the Association for
               Computational Linguistics (ACL-IJCNLP)},
  pages     = {3277--3287},
  year      = {2021}
}

@article{convfinqa,
  title   = {ConvFinQA: Exploring the Chain of Numerical Reasoning in Conversational
             Finance Question Answering},
  author  = {Chen, Zhiyu and Li, Shiyang and Smiley, Jamie and Ma, Zhiqiang and
             Shah, Meghana Bhat and Shah, Chitta and Yang, Zewei and Shi, Heng and
             Wang, William Yang},
  journal = {arXiv preprint arXiv:2210.03849},
  year    = {2022}
}

@article{chen2021finqa,
  title   = {FinQA: A Dataset of Numerical Reasoning over Financial Data},
  author  = {Chen, Zhiyu and Chen, Wenhu and Smiley, Charese and Shah, Sameena and Borova, Iana and Langdon, Dylan and Moussa, Reema and Beane, Matt and Huang, Ting-Hao and Routledge, Bryan and Wang, William Yang},
  journal = {arXiv preprint arXiv:2109.00122},
  year    = {2021}
}

@article{llm_limits_math,
  title   = {Training Verifiers to Solve Math Word Problems},
  author  = {Cobbe, Karl and Kosaraju, Vineet and Bavarian, Mohammad and Chen, Mark
             and Jun, Heewoo and Kaiser, Lukasz and Tworek, Jerry and Hilton, Jacob
             and Nakano, Reiichiro and Hesse, Christopher and Schulman, John},
  journal = {arXiv preprint arXiv:2110.14168},
  year    = {2021}
}

@article{gsm_symbolic,
  title   = {{GSM-Symbolic}: Understanding the Limitations of Mathematical Reasoning
             in Large Language Models},
  author  = {Mirzadeh, Iman and Alizadeh-Vahid, Keivan and Shahrokhi, Hooman and
             Tuzel, Oncel and Bengio, Samy and Farajtabar, Mehrdad},
  journal = {arXiv preprint arXiv:2410.05229},
  year    = {2024}
}

@article{numerical_precision_llm,
  title   = {How Numerical Precision Affects Mathematical Reasoning Capabilities of
             {LLMs}},
  author  = {Yang, Guhao and Wen, Zhen and Chen, Jie and Hu, Jianzhun and
             Du, Simon S. and Wang, Liwei},
  journal = {arXiv preprint arXiv:2410.13857},
  year    = {2024}
}

@article{llm_math_errors,
  title   = {Mathematical Reasoning in Large Language Models: Assessing Logical and
             Arithmetic Errors across Wide Numerical Ranges},
  author  = {Singh, Aditya and Bhattamishra, Satwik and Bhattacharyya, Pushpak},
  journal = {arXiv preprint arXiv:2502.08680},
  year    = {2025}
}

@article{dziri_compositionality,
  title   = {Faith and Fate: Limits of Transformers on Compositionality},
  author  = {Dziri, Nouha and Lu, Ximing and Sclar, Melanie and Li, Xiang Lorraine
             and Jian, Liwei and Lin, Bill Yuchen and West, Peter and Bhagavatula,
             Chandra and Bras, Ronan Le and Hwang, Jena D. and Sanyal, Soumya and
             Welleck, Sean and Bhatt, Gagan and Ruder, Sebastian and Ren, Xiang
             and Ettinger, Allyson and Harchaoui, Zaid and Choi, Yejin},
  journal = {arXiv preprint arXiv:2305.18654},
  year    = {2023}
}

@inproceedings{rag,
  title     = {Retrieval-Augmented Generation for Knowledge-Intensive {NLP} Tasks},
  author    = {Lewis, Patrick and Perez, Ethan and Piktus, Aleksandra and Petroni,
               Fabio and Karpukhin, Vladimir and Goyal, Naman and K{\"u}ttler,
               Heinrich and Lewis, Mike and Yih, Wen-tau and Rockt{\"a}schel, Tim
               and others},
  booktitle = {Advances in Neural Information Processing Systems (NeurIPS)},
  year      = {2020}
}

@article{rag_survey,
  title   = {Retrieval-Augmented Generation for Large Language Models: A Survey},
  author  = {Gao, Yunfan and Xiong, Yun and Gao, Xinyu and Jia, Kangxiang and
             Pan, Jinliu and Bi, Yuxi and Dai, Yi and Sun, Jiawei and Wang, Haofen},
  journal = {arXiv preprint arXiv:2312.10997},
  year    = {2023}
}

@article{self_rag,
  title   = {Self-{RAG}: Learning to Retrieve, Generate, and Critique through
             Self-Reflection},
  author  = {Asai, Akari and Wu, Zeqiu and Wang, Yizhong and Sil, Avirup and
             Hajishirzi, Hannaneh},
  journal = {arXiv preprint arXiv:2310.11511},
  year    = {2023}
}

@inproceedings{pal,
  title     = {Program-Aided Language Models},
  author    = {Gao, Luyu and Madaan, Aman and Zhou, Shuyan and Alon, Uri and
               Yang, Pengfei and Callan, Jamie and Neubig, Graham},
  booktitle = {International Conference on Machine Learning (ICML)},
  year      = {2023}
}

@inproceedings{toolformer,
  title     = {Toolformer: Language Models Can Teach Themselves to Use Tools},
  author    = {Schick, Timo and Dwivedi-Yu, Jane and Dess{\`\i}, Roberto and
               Raileanu, Roberta and Lomeli, Maria and Hambro, Eric and
               Zettlemoyer, Luke and Cancedda, Nicola and Scialom, Thomas},
  booktitle = {Advances in Neural Information Processing Systems (NeurIPS)},
  year      = {2023}
}

@inproceedings{react,
  title     = {{ReAct}: Synergizing Reasoning and Acting in Language Models},
  author    = {Yao, Shunyu and Zhao, Jeffrey and Yu, Dian and Du, Nan and
               Shafran, Izhak and Narasimhan, Karthik and Cao, Yuan},
  booktitle = {International Conference on Learning Representations (ICLR)},
  year      = {2023}
}

@article{gorilla,
  title   = {Gorilla: Large Language Model Connected with Massive {APIs}},
  author  = {Patil, Shishir G. and Zhang, Tianjun and Wang, Xin and Gonzalez,
             Joseph E.},
  journal = {arXiv preprint arXiv:2305.15334},
  year    = {2023}
}

@article{api_bank,
  title   = {{API-Bank}: A Comprehensive Benchmark for Tool-Augmented {LLMs}},
  author  = {Li, Minghao and Zhao, Yingxiu and Yu, Bowen and Song, Feifan and
             Li, Hangyu and Yu, Haiyang and Li, Zhoujun and Huang, Fei and
             Li, Yongbin},
  journal = {arXiv preprint arXiv:2304.08244},
  year    = {2023}
}

@inproceedings{pot,
  title     = {Program of Thoughts Prompting: Disentangling Computation from
               Reasoning for Numerical Reasoning Tasks},
  author    = {Chen, Wenhu and Ma, Xueguang and Wang, Xinyi and Cohen, William W.},
  booktitle = {Transactions on Machine Learning Research},
  year      = {2023}
}

@inproceedings{cot,
  title     = {Chain-of-Thought Prompting Elicits Reasoning in Large Language
               Models},
  author    = {Wei, Jason and Wang, Xuezhi and Schuurmans, Dale and Bosma, Maarten
               and Chi, Ed and Narang, Sharan and Chowdhery, Aakanksha and Le, Quoc
               and Zhou, Denny},
  booktitle = {Advances in Neural Information Processing Systems (NeurIPS)},
  year      = {2022}
}

@article{least_to_most,
  title   = {Least-to-Most Prompting Enables Complex Reasoning in Large Language
             Models},
  author  = {Zhou, Denny and Sch{\"a}rli, Nathanael and Hou, Le and Wei, Jason
             and Scales, Nathan and Wang, Xuezhi and Schuurmans, Dale and
             Cui, Claire and Bousquet, Olivier and Le, Quoc and Chi, Ed},
  journal = {arXiv preprint arXiv:2205.10625},
  year    = {2022}
}

@article{self_consistency,
  title   = {Self-Consistency Improves Chain of Thought Reasoning in Language
             Models},
  author  = {Wang, Xuezhi and Wei, Jason and Schuurmans, Dale and Le, Quoc and
             Chi, Ed and Narang, Sharan and Chowdhery, Aakanksha and Zhou, Denny},
  journal = {arXiv preprint arXiv:2203.11171},
  year    = {2022}
}

@inproceedings{tot,
  title     = {Tree of Thoughts: Deliberate Problem Solving with Large Language
               Models},
  author    = {Yao, Shunyu and Yu, Dian and Zhao, Jeffrey and Shafran, Izhak and
               Griffiths, Thomas L. and Cao, Yuan and Narasimhan, Karthik},
  booktitle = {Advances in Neural Information Processing Systems (NeurIPS)},
  year      = {2023}
}

@article{autogen,
  title   = {{AutoGen}: Enabling Next-Gen {LLM} Applications via Multi-Agent
             Conversation},
  author  = {Wu, Qingyun and Bansal, Gagan and Zhang, Jieyu and Wu, Yiran and
             Li, Beibin and Zhu, Erkang and Jiang, Li and Zhang, Xiaoyun and
             Liu, Chi and Awadallah, Ahmed and others},
  journal = {arXiv preprint arXiv:2308.08155},
  year    = {2023}
}

@article{camel,
  title   = {{CAMEL}: Communicative Agents for Mind Exploration of Large Language
             Model Society},
  author  = {Li, Guohao and Hammoud, Hasan Abed Al Kader and Itani, Hani and
             Khizbullin, Dmitrii and Ghanem, Bernard},
  journal = {arXiv preprint arXiv:2303.17760},
  year    = {2023}
}

@article{metagpt,
  title   = {{MetaGPT}: Meta Programming for Multi-Agent Collaborative Frameworks},
  author  = {Hong, Sirui and Zhuge, Ming and Chen, Jonathan and Zheng, Xiaoxuan
             and Cheng, Yulin and Zhang, Chaoyi and Wang, Jian and Wang, Zhonghao
             and Yau, Shing-Chi and Lin, Zhiqing and others},
  journal = {arXiv preprint arXiv:2308.00352},
  year    = {2023}
}

@article{multiagent_llm,
  title   = {Multi-Agent Collaboration with Large Language Models},
  author  = {Wang, Xuezhi and others},
  journal = {arXiv preprint arXiv:2304.03442},
  year    = {2023}
}

@inproceedings{reflexion,
  title     = {Reflexion: Language Agents with Verbal Reinforcement Learning},
  author    = {Shinn, Noah and Cassano, Federico and Berman, Edward and
               Gopinath, Ashwin and Narasimhan, Karthik and Yao, Shunyu},
  booktitle = {Advances in Neural Information Processing Systems (NeurIPS)},
  year      = {2023}
}

@article{agent_survey,
  title   = {A Survey on Large Language Model Based Autonomous Agents},
  author  = {Wang, Lei and Ma, Chen and Feng, Xueyang and Zhang, Zeyu and Yang,
             Hao and Zhang, Jingsen and Chen, Zhiyuan and Tang, Jiakai and Chen,
             Xu and Lin, Yankai and Zhao, Wayne Xin and Wei, Zhewei and Wen,
             Ji-Rong},
  journal = {Frontiers of Computer Science},
  volume  = {18},
  number  = {6},
  pages   = {186345},
  year    = {2024}
}

@article{llm_agent_planning_survey,
  title   = {Understanding the Planning of {LLM} Agents: A Survey},
  author  = {Huang, Xu and Liu, Weiwen and Chen, Xiaolong and Wang, Xingmei and
             Wang, Hao and Lian, Defu and Wang, Yasheng and Tang, Ruiming and
             Chen, Enhong},
  journal = {arXiv preprint arXiv:2402.02716},
  year    = {2024}
}

@article{neurosymbolic,
  title   = {Neuro-Symbolic {AI}: The Third Wave},
  author  = {Garcez, Artur d'Avila and Lamb, Luis C.},
  journal = {AI Magazine},
  volume  = {41},
  number  = {2},
  pages   = {21--36},
  year    = {2020}
}

@article{neurosymbolic_review_2025,
  title   = {A Review of Neuro-Symbolic {AI} Integrating Reasoning and Learning
             for Advanced Cognitive Systems},
  author  = {Nawaz, Uzma and Anees-ur-Rahaman, Mufti and Saeed, Zubair},
  journal = {Intelligent Systems with Applications},
  volume  = {26},
  pages   = {200541},
  year    = {2025}
}

@article{llm_symbolic_limits,
  title   = {Architectural Limits of {LLMs} in Symbolic Computation and Structured
             Reasoning},
  author  = {Charton, François and others},
  journal = {arXiv preprint arXiv:2507.10624},
  year    = {2025}
}

@article{llama2,
  title   = {{Llama} 2: Open Foundation and Fine-Tuned Chat Models},
  author  = {Touvron, Hugo and Martin, Louis and Stone, Kevin and Albert, Peter and
             Almahairi, Amjad and Babaei, Yasmine and Bashlykov, Nikolay and
             Batra, Soumya and Bhargava, Prajjwal and Bhosale, Shruti and others},
  journal = {arXiv preprint arXiv:2307.09288},
  year    = {2023}
}

@article{llama3,
  title   = {The {Llama} 3 Herd of Models},
  author  = {Dubey, Abhimanyu and Jauhri, Abhinav and Pandey, Abhinav and Kadian,
             Abhishek and Al-Dahle, Ahmad and Letman, Aiesha and Mathur, Akhil
             and others},
  journal = {arXiv preprint arXiv:2407.21783},
  year    = {2024}
}

@article{scaling_laws,
  title   = {Scaling Laws for Neural Language Models},
  author  = {Kaplan, Jared and McCandlish, Sam and Henighan, Tom and Brown,
             Tom B. and Chess, Benjamin and Child, Rewon and Gray, Scott and
             Radford, Alec and Wu, Jeffrey and Amodei, Dario},
  journal = {arXiv preprint arXiv:2001.08361},
  year    = {2020}
}

@article{emergent_abilities,
  title   = {Emergent Abilities of Large Language Models},
  author  = {Wei, Jason and Tay, Yi and Bommasani, Rishi and Raffel, Colin and
             Zoph, Barret and Borgeaud, Sebastian and Yogatama, Dani and
             Bosma, Maarten and Zhou, Denny and Metzler, Donald and Chi, Ed H.
             and Hashimoto, Tatsunori and Vinyals, Oriol and Liang, Percy and
             Dean, Jeff and Fedus, William},
  journal = {Transactions on Machine Learning Research},
  year    = {2022}
}

@article{hallucination_survey,
  title   = {A Survey on Hallucination in Large Language Models: Principles,
             Taxonomy, Challenges, and Open Questions},
  author  = {Huang, Lei and Yu, Weijiang and Ma, Weitao and Zhong, Weihong and
             Feng, Zhangyin and Wang, Haotian and Chen, Qianglong and Peng, Weihua
             and Feng, Xiaocheng and Qin, Bing and Liu, Ting},
  journal = {arXiv preprint arXiv:2311.05232},
  year    = {2023}
}

@article{factuality_llm,
  title   = {Survey of Hallucination in Natural Language Generation},
  author  = {Ji, Ziwei and Lee, Nayeon and Frieske, Rita and Yu, Tiezheng and
             Su, Dan and Xu, Yan and Zeng, Aixin and Fung, Yue-En Suh and
             Hong, Yuxiang and Fung, Pascale},
  journal = {ACM Computing Surveys},
  volume  = {55},
  number  = {12},
  pages   = {1--38},
  year    = {2023}
}

\end{document}